\documentclass[10pt,twocolumn,letterpaper]{article}

\usepackage{cvpr}
\usepackage{times}
\usepackage{epsfig}
\usepackage{graphicx}

\usepackage{algorithmic}
\usepackage{multirow}

\usepackage{color, colortbl}

\usepackage[ruled,vlined,linesnumbered]{algorithm2e}
\usepackage{array}
\usepackage[english]{babel} 
\usepackage{amsmath,amsfonts,amsthm} 
\usepackage{ragged2e}

\usepackage{makecell}

\usepackage{authblk}

\usepackage[breaklinks=true,bookmarks=false]{hyperref}

\cvprfinalcopy 

\def\cvprPaperID{594} 

\begin{document}

\title{From Zero-shot Learning to Conventional Supervised Classification: Unseen Visual Data Synthesis}

\author[1]{Yang Long}
\author[2]{Li Liu}
\author[2]{Ling Shao}
\author[3]{Fumin Shen}
\author[4]{Guiguang Ding}
\author[5]{Jungong Han}

\affil[1]{Department of Electronic and Electrical Engineering, University of Sheffield, UK}
\affil[2]{School of Computing Sciences, University of East Anglia, UK}
\affil[3]{Center for Future Media, University of Electronic Science and Technology of China, China}
\affil[4]{School of Software, Tsinghua University, China}
\affil[5]{Department of Computer Science and Digital Technologies, Northumbria University, UK}
\affil{\small{\texttt{ylong2@sheffield.ac.uk, \{li.liu, ling.shao\}@uea.ac.uk, fumin.shen@gmail.com, dinggg@tsinghua.edu.cn, jungong.han@northumbria.ac.uk}}}

%
%

\maketitle

\begin{abstract}
	Robust object recognition systems usually rely on powerful feature extraction mechanisms from a large number of real images. However, in many realistic applications, collecting sufficient images for ever-growing new classes is unattainable. In this paper, we propose a new Zero-shot learning (ZSL) framework that can synthesise visual features for unseen classes without acquiring real images. Using the proposed \textit{Unseen Visual Data Synthesis} (UVDS) algorithm, semantic attributes are effectively utilised as an intermediate clue to synthesise unseen visual features at the training stage. Hereafter, ZSL recognition is converted into the conventional supervised problem, \ie the synthesised visual features can be straightforwardly fed to typical classifiers such as SVM. On four benchmark datasets, we demonstrate the benefit of using synthesised unseen data. Extensive experimental results suggest that our proposed approach significantly improve the state-of-the-art results.
\end{abstract}

\section{Introduction}
Object Recognition is arguably one of the most fundamental tasks in computer vision field. Most of the conventional frameworks, \eg Deep Neural Networks (DNN) \cite{CNN},  rely on a large number of training samples to build statistical models. However, such a premise is unattainable in many real-world situations. The main reasons can be summarised as follows: 1) Obtaining well-annotated training samples is expensive. Although abundant digital images and videos can be retrieved from the Internet, existing search engines crucially depend on user-defined keywords that are often vague and not suitable for learning tasks. 2) The number of newly defined classes is ever-growing. Meanwhile, fine-grained tasks make existing categories go deeper, \eg to recognise a newly released handbag in a novel pattern. Training a particular model for each of them is infeasible. 3) Collecting instances for rare classes is difficult. For example, one might wish to detect an ancient or rare species automatically. It could be difficult to provide even a single example for them since available knowledge could be only textual descriptions or some distinctive attributes.
\begin{figure}
	\centering
	\includegraphics[width=0.4\textwidth]{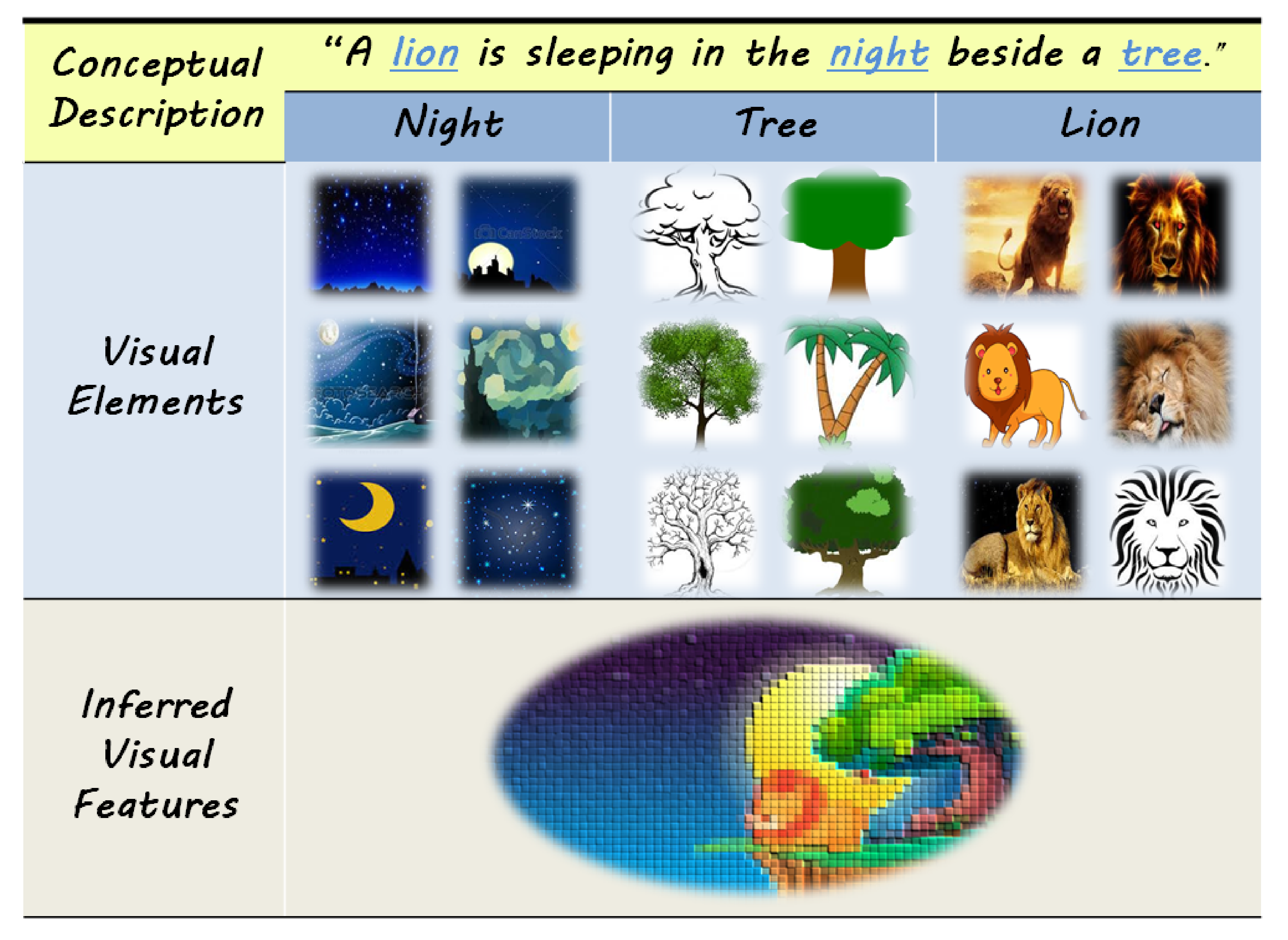}
	\caption{Given a conceptual description, human can imagine the outline of the scene by combining previous seen visual elements.}
	\label{Fig0_illu}
	\vspace{-2ex}
\end{figure}

\begin{figure*}
	\centering
	\includegraphics[width=16cm]{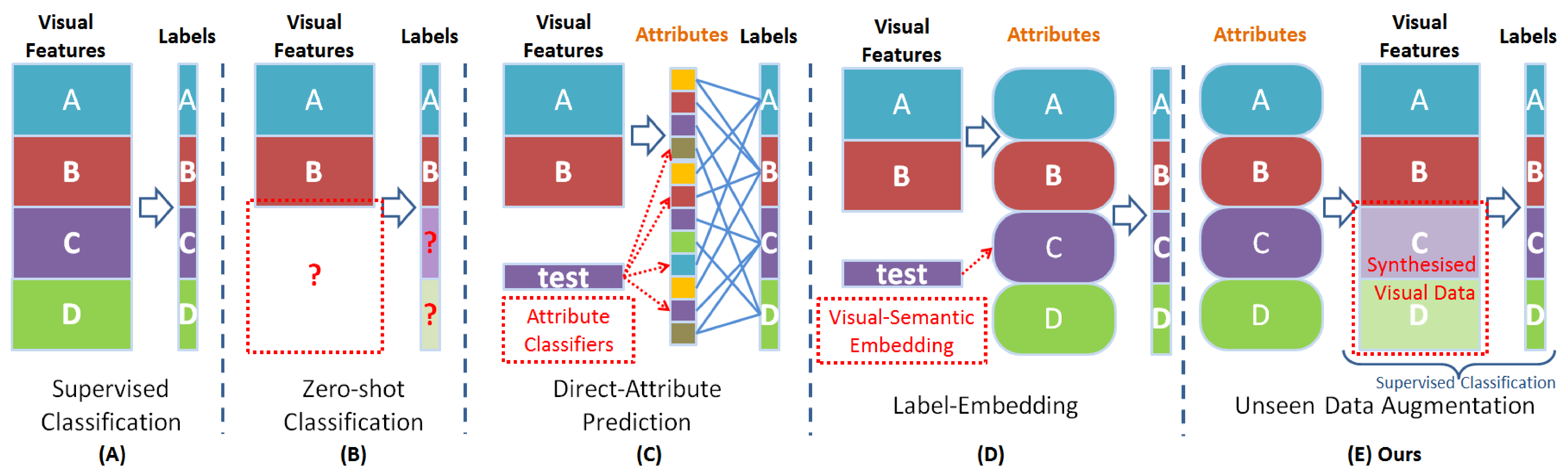}
	\caption{\small Comparison of supervised and zero-shot classifications and existing ZSL frameworks. (A) a typical supervised classification: the training samples and labels are in pairs; (B) a zero-shot learning problem: without training samples, the classes \textit{C} and \textit{D} cannot be predicted; (C) Direct-Attribute Prediction model uses attributes as intermediate clues to associate visual features to class labels; (D) label-embedding: the attributes are concatenated as a semantic embedding; (E) we inversely learn an embedding from the semantic space to visual space and convert the ZSL problem into conventional supervised classification.}
	\label{Fig_frame_cmp}
	\vspace{-2ex}
\end{figure*}

As a feasible solution, \textit{Zero-shot Learning} (ZSL) aims to leverage a closed-set of semantic models that can generalise to unseen classes \cite{2008_zero_data,2009_AwA}. The common paradigm of ZSL methods first train a prediction model that can map visual data to a semantic representation. Hereafter, new objects can be recognised by only knowing their semantic descriptions. However, existing methods cannot expand the training data for new unseen classes. As illustrated in Fig. \ref{Fig_frame_cmp}, such frameworks impede existing methods from scaling up since the fixed seen data is eventually limited to represent the ever-growing semantic concepts.

In this paper, we investigate to synthesise high-quality visual features from semantic attributes so that the ZSL problem can be converted into conventional supervised classification. Our idea is inspired by the ability of human imagination, as shown in Fig. \ref{Fig0_illu}. Given a semantic description, we human can associate familiar visual elements and then imagine an approximate scene. Accordingly, we synthesise discriminative low-level features from semantic attributes to substitute feature extraction from real images. Our contributions can be summarised as follows:

\textbf{1) }We provide a feasible framework to synthesise unseen visual features from given semantic attributes without acquiring real images. The synthesised data obtained at the training stage can be straightforwardly fed to conventional classifiers so that ZSL recognition is skilfully converted into the conventional supervised problem and leads to state-of-the-art recognition performance on four benchmark datasets.

\textbf{2)} We introduce the \textit{variance decay} problem during semantic-visual embedding and propose a novel \textit{Diffusion Regularisation} that can explicitly make information diffuse to each dimension of the synthesised data. We achieve information diffusion by optimising an orthogonal rotation problem. We provide an efficient optimisation strategy to solve this problem together with the \textit{structural difference} and \textit{training bias} problem.


\section{Related Work}
\noindent\textbf{Zero-shot Recognition Schemes:} We summarise previous ZSL schemes in Fig. \ref{Fig_frame_cmp}, in contrast to conventional supervised classification (Fig. \ref{Fig_frame_cmp}(A)). Since collecting well-labelled visual data for novel classes is expensive, as shown in Fig. \ref{Fig_frame_cmp}(B), zero-shot learning techniques \cite{2008_zero_data,2009_AwA,add5,2009_Hinton,add1,add2} are proposed to recognise novel classes without acquiring the visual data. Most of the early works are based on the Direct-Attribute Prediction (DAP) model \cite{2009_AwA}. Such a model utilises semantic attributes as intermediate clues. A test sample is classified by each attribute classifier alternately, and the class label is predicted by probabilistic estimation. Admitting the merit of DAP, there are some concerns about its deficiencies. \cite{2014_Decorrelating} points out that the attributes may correlate to each other resulting in significant information redundancy and poor performance. The human labelling involved in attribute annotation may also be unreliable \cite{2014_Unreliable,add6}.

To circumvent learning independent attributes, embedding-based ZSL frameworks (Fig.\ref{Fig_frame_cmp}(C)) are proposed to learn a projection that can map the visual features to all of the attributes at once.  The class label is then inferred in the semantic space using various measurements \cite{2013_Label_Embedding,2012_Metric_Learning,2015_Max_Margin,2014_LatentAttributeSpace,2015_Semantic_Manifold,2013_crossModal_Ng}. Since the attribute annotations are expansive to acquire, attributes are substituted by the visual similarity and data distribution information in transductive ZSL settings \cite{2013_NIPS_transductive,2015_ICLR_Hospedales,2015_PAMI_GsG_Multiview,2014_ECCV_GSG_Multiview,2015_Semi_supervised,2015_unsupervised_domain,2015_semantic_similarity,2016_ZZiming,2016_ECCV_ZZming}. However, these methods involve the data of unseen classes to learn the model, which to some extent breaches the strict ZSL settings. Recent work \cite{2015_embarrassingly,2016_CVPR_latent_embedding,add4} combines the embedding-inferring procedure into a unified framework and empirically demonstrates better performance. The closest related work is \cite{2016_synthesized_Classifiers,2016_arXiv_predictingExemplar,add3}, which takes one-step further to synthesise classifiers or prototypes for unseen classes.

Our method takes the advantages of semantic embedding. However, the inference direction is different from existing work. Our method aims to inversely synthesise visual feature vectors to as many as the available semantic instances rather than mapping visual data to the label space.

\noindent\textbf{Semantic Side Information:} ZSL tasks require to leverage side information as intermediate clues.  Such frameworks not only broaden the classification settings but also enable various information to aid visual systems. Since textual sources are relatively easy to obtain from the Internet, \cite{2010_What_helps,2014_COSTA} propose to estimate the semantic relatedness of the novel classes from the text. \cite{2015_Deep_ZS_Textual,2013_Write_A_Classifier,2015_Deep_ZS_Textual} learn pseudo-concepts to associate novel classes using Wikipedia articles. Recently, lexical hierarchies in the ontology engineering are also exploited to find the relationships between classes \cite{2011_evaluate_largescale,2015_WACV,2015_akata_evaluation}.

Although various side information is studied, attribute-based ZSL methods still gain the most popularity. One reason is ZSL by learning attributes often gives prominent classification performance \cite{2010_One_or_Zero,2013_Category_level,2015_Hypergraph,2016_ZZiming,2015_semantic_similarity}. For another reason, attribute representation is a compact way that can further describe an image by concrete words that are human-understandable \cite{2009_aPY,2015_Multiplicative,2016_AAAI_Not_Alone,2016_akata_Strong_Supervision}. Various types of attributes are proposed to enrich applicable tasks and improve the performance, such as relative attributes \cite{2011_relative}, class-similarity attributes \cite{2013_Category_level}, and augmented attributes \cite{2012_augmented}. Our main motivation of this paper not only aims to improve the ZSL performance, but also seeks for a reliable solution for synthesising high-quality visual features.

\section{Approach}
\noindent\textbf{Preliminaries} The training set contains \textit{centralised} visual features, attributes, and seen class labels that are in 3-tuples: $(x_1,a_1,y_1),...,(x_N,a_N,y_N) \subseteq \mathcal{X}_s \times\mathcal{A}_s \times\mathcal{Y}_s$, where $N$ is the number of training samples; $\mathcal{X}_s=[x_{nd}]\in\mathbb{R}^{N\times D}$ is a $D$-dimensional feature space; $\mathcal{A}_s=[a_{nm}]\in\mathbb{R}^{N\times M}$ is an $M$-dimensional attribute space; and $y_n \in \{1,...,C\}$ consists of $C$ discrete class labels. Our framework can cope with either \textit{class-level} or \textit{image-level} attributes. For class-level, the instances in the same class share the attributes. Given $\hat{N}$ pairs of instances with semantic attributes from $\hat{C}$ unseen classes: $(\hat{a}_1,\hat{y}_1),...,(\hat{a}_{\hat{N}},\hat{y}_{\hat{N}}) \subseteq \mathcal{A}_u \times\mathcal{Y}_u$, where $\mathcal{Y}_u\cap\mathcal{Y}_s= \emptyset$, $\mathcal{A}_u=[a_{\hat{n}m}]\in\mathbb{R}^{\hat{N}\times M}$, the goal of zero-shot learning is to learn a classifier, $f: \mathcal{X}_u \rightarrow \mathcal{Y}_u$, where the samples in $\mathcal{X}_u$ are completely unavailable during training.
We use \textit{Calligraphic} typeface to indicate a space. Subscripts $s$ and $u$ refer to `seen' and `unseen'. \textit{hat} denotes the variables that are related to `unseen' samples.

\noindent\textbf{Unseen Visual Data Synthesis:} We aim to synthesise the visual features of unseen classes by the given semantic attributes. Specifically, we learn an embedding function on the training set $f': \mathcal{A}_s \rightarrow \mathcal{X}_s$. After that, we are able to infer $\mathcal{X}_u$ through: $\mathcal{X}_u = f'(\mathcal{A}_u)$.

\noindent\textbf{Zero-shot Recognition:} Using the synthesised visual features, the ZSL recognition is converted to a typical classification problem. It is straightforward to employ conventional supervised classifiers, \eg SVM, to predict the labels of unseen classes $f_{\text{SVM}}: \mathcal{X}_u \rightarrow \mathcal{Y}_u$.

\subsection{Unseen Visual Data Synthesis}
To synthesise visual features, the most intuitive framework is to learn a mapping function from the semantic space to the visual feature space:
\begin{equation}
	\min_P\mathcal{L}(\mathcal{A}_sP,\mathcal{X}_s)+\lambda \Omega (P) \text{,}
	\label{f2_A2X}
\end{equation}
where $P$ is the projection matrix, $\mathcal{L}$ is a loss function, and $\Omega$ is a regularisation term with its hyper-parameter $\lambda$. It is common to choose $\Omega (P) = \| P \|_F^2$, where $\|.\|_F$ is the Frobenius norm of a matrix that estimates the Euclidean distance between two matrices.  Before the test, we can synthesise unseen visual features from the attribute space by given attributes of the unseen instances:
\begin{equation}
	\mathcal{X}_u=\mathcal{A}_uP.
	\label{f3_XWA}
\end{equation}

\noindent\textbf{Visual-Semantic Structure Preservation}
In spite of the simplicity of the above framework, we confront two main problems as follows. 1) \textit{Structural difference}: in practice, there is often a huge gap between visual and semantic spaces. In pursuance of minimum reconstruction error, the model tends to learn principal components between the two spaces. Consequently, the synthesised data would be not discriminant enough for ZSL purposes. 2) \textit{Training bias}: the synthesised unseen data can be biased towards the `seen' data and gains a different data distribution to the real unseen data. This problem is due to the regression-based framework does not discover the intrinsic geometric structure of the semantic space and cannot capture the unseen-to-seen relationships. Thus, directly mapping from semantic to visual space can lead to inferior performance. We propose to introduce an auxiliary latent-embedding space $\mathcal{V}$ to reconcile the semantic space with the visual feature space, where  $\mathcal{V}=[v_{nd}]\in\mathbb{R}^{N\times D}$. In this way, instead of $\Omega (P)$, we can let $\mathcal{V}$ preserve the intrinsic data structural information of both visual and semantic spaces:
\begin{equation}
	J= \|\mathcal{X}_s-\mathcal{V}Q\|^2_F+\|\mathcal{V}-\mathcal{A}_s P\|^2_F +\lambda\Omega_1(\mathcal{V}) \text{,}
	\label{f3_XVA+O(V)}
\end{equation}
where the latent-embedding space $\mathcal{V}$ is decomposed from $\mathcal{X}$ and $\mathcal{A}$ is then decomposed from $\mathcal{V}$. $Q=[q_{d'd}]\in\mathbb{R}^{D\times D}$ and $P=[p_{md}]\in\mathbb{R}^{M\times D}$ are two projection matrices. $\Omega_1$ is a \textit{dual-graph} that is introduced next.

We take the \textit{Local Invariance} \cite{NMF} assumption and solve the problem through a spectral \textit{Dual-Graph} approach. This is a combination of two supervised graphs that aim to simultaneously estimate the data structures of both $\mathcal{X}$ and $\mathcal{A}$. The graph of visual space $W_\mathcal{X}\in\mathbb{R}^{N\times N}$ has $N$ vertices $\{g_1,...,g_N\}$ that correspond to $N$ data points $\{x_1,...,x_N\}$ in the training set. The semantic graph $W_\mathcal{A}\in\mathbb{R}^{N\times N}$ has the same number of vertices as $N$ instances of attributes $\{a_1,...,a_N\}$.  For \textit{image-level attributes}, we construct $k$-nn graphs for both visual and semantic spaces, \ie put an edge between each data point $x_n$ (or $a_n$) and each of its $k$ nearest neighbours.  For each pair of the vertices $g_i ~\text{and}~ g_j$ in the weight matrix (not differ in $W_\mathcal{X}$ and $W_\mathcal{A}$), the weight can be defined as
\begin{equation}\label{e2}
	w_{ij} =\left\{
	\begin{tabular}{ll}
		1, & if\ \ $g_i ~\text{and}~ g_j$ are connected by an edge\\
		0, & otherwise.
	\end{tabular}\right.
\end{equation}
As a result, we can separately compute the two weight matrices $W_\mathcal{X}$ and $W_\mathcal{A}$. It is noteworthy that, for \textit{class-level attributes}, $W_\mathcal{A}$ is computed in a slightly different way. Every vertex in the same class is connected by a normalised edge, i.e. $w_{ij}= k\slash n_c$, if and only if $a_{i} ~\text{and}~ a_{j}$ are from the same class $c$, where $n_c$ is the size of class $c$.

In the embedding space $\mathcal{V}$, we expect that, if $g_i$ and $g_j$ in both graphs are connected, each pair of embedded points $v_{i}$ and $v_{j}$ are also close to each other. However, sometimes $W_{\mathcal{X}}$ and $W_{\mathcal{A}}$ are not always consistent due to the visual-semantic gap. To compromise such conflicts, we compute the mean of the visual and attribute graphs, i.e. $W=\frac{1}{2}(W_{\mathcal{X}}+ W_{\mathcal{A}})$. The resulted regularisation is:
\begin{equation}
	\begin{aligned}
		\Omega_1 (\mathcal{V})&=\frac{1}{2} \displaystyle{\sum_{i,j=1}^{N}}\|v_{i}-v_{j}\|^2w_{ij}\\
		&=Tr(\mathcal{V}^\top\mathbf{D}\mathcal{V})-Tr(\mathcal{V}^\top W\mathcal{V})=Tr(\mathcal{V}^\top L\mathcal{V}) \text{,}
	\end{aligned}
	\label{f2_L-DW}
\end{equation}
where $\mathbf{D}$ is the degree matrix of $W$, $\mathbf{D}_{ii}=\sum_{i}w_{ij}$. $L$ is known as graph Laplacian matrix $L=\mathbf{D}-W$ and $Tr(.)$ computes the trace of a matrix.

\noindent\textbf{Diffusion Regularisation}
In this paper, we identify another fundamental problem: \textit{variance decay}.
When we learn visual features from the attributes, in particular when projecting $\mathcal{A}$ to $\mathcal{V}$ using $\mathcal{P}$, the dimension difference $D\gg M$ will lead the learning algorithm to pick the directions with low variances progressively. As shown in Fig. \ref{fig_DRcurve}, most of the information (variance) is contained in a few projections. As a result, the remaining dimensions of the synthesised data suffers a dramatic variance decay, which indicates the learnt representation is severely redundant. To address the problem, we may expect the concentrated information can effectively diffuse to all of the learnt dimensions through an adjustment rotation \cite{Rotation}. Therefore, we modify the rotating matrix $Q$ in Eq. (\ref{f3_XVA+O(V)}). In this paper, we consider an orthogonal rotation, \ie $QQ^\top=I$, since it is easy to show that $Tr(Q^\top P^\top\mathcal{A}^\top\mathcal{A}PQ) = Tr(P^\top\mathcal{A}^\top\mathcal{A}P)$ ($I$ is an identity matrix). Such a property is reported in \cite{Protect} that the orthogonal rotation can protect the properties captured in the semantic space. Next, we show how the rotation can control variance diffusion.

From Eq. (\ref{f3_XVA+O(V)}), the optimal synthesised data is $\mathcal{X}=\mathcal{V}Q$, where $\mathcal{V}=\mathcal{A}P$. We first prove that the overall variance does not change after rotation. Before rotation, $\mathcal{V}$ is centralised, i.e. $\sum_{n=1}^{N} v_n = \textbf{0}$.  The original overall variance $\Gamma$ of $\mathcal{V}$ is $\Gamma = N\sum_{d=1}^{D}\sigma_d$, where $\sigma_d=(\sum_{n=1}^{N} v^2_{nd})/N$ denotes the variance of the $d$-th dimension. After rotation $Q$, we have the new variance of each dimension $\sigma_d'$ and the sum of variance of each dimension is $\Gamma'$. We show $\Gamma = \Gamma'$ in the following:
\begin{eqnarray}
	\Gamma &=& 
	\sum_{d=1}^{D}\sum_{n=1}^{N} v_{nd}^2 = \|\mathcal{V}\|_F^2 =Tr(\mathcal{V}\mathcal{V}^\top) \nonumber \\
	&=& Tr(\mathcal{V}QQ^\top\mathcal{V}^\top) = \|\mathcal{V}Q\|_F^2 \nonumber \\
	&=& \sum_{d=1}^{D}\sum_{n=1}^{N} {x}_{nd}^2 = N\sum_{d=1}^{D}\sigma'_d=\Gamma'\text{.}
	\label{theorem1_invariance}
\end{eqnarray}

We hope the overall variance $\Gamma$ tends to equally diffuse to all of the learnt dimensions in order to recover the real data distribution of $\mathcal{X}$. In other words, the variance of diffused standard deviations $\Pi$ in the synthesised data should be small ($\Pi =  \frac{1}{D}\sum_{d=1}^{D} (\pi_d - \bar{\pi})^2$, where $\pi_d=\sqrt{\sigma'_d} $ and $\bar{\pi}$ is the mean of all standard deviations). According to the above Eq. (\ref{theorem1_invariance}), we have $\sum_{d=1}^{D} \pi_{d}^2 = \sum_{d=1}^{D} {\sigma'_{d}} = \sum_{d=1}^{D} \sigma_{d}= \epsilon$. Next, we show how to minimise $\Pi$ in our learning framework to find the orthogonal rotation:
\begin{eqnarray}
	\Pi &=& \frac{1}{D}\sum_{d=1}^{D}(\pi_d-\bar{\pi})^2 \nonumber \\
	&=& \frac{1}{D}\sum_{d=1}^{D}\pi_d^2 + \bar{\pi}^2 - \frac{2}{D}\sum_{d=1}^{D}\pi_d\bar{\pi} \nonumber \\
	&=& \frac{\epsilon}{D}-\frac{1}{D^2}(\sum_{d=1}^{D} \pi_d)^2 \text{.}
	\label{Eq_bigPi}
\end{eqnarray}

The above equation shows that to minimise $\Pi$ is equivalent to maximise the sum of diffused standard deviations. Such a deduction is intuitive because our goal is a higher overall sum of standard deviation so that the synthesised data can gain more information. Moreover, we discover a novel relationship between the sum of diffused standard deviations and the orthogonal rotation:
\begin{eqnarray}
	\sum_{d=1}^{D} \pi_d &=& \sum_{d=1}^{D} \sqrt{\sigma'_d} = \sum_{d=1}^{D} \sqrt{\sum_{n=1}^{N} x^2_{nd}/N} \nonumber\\
	&=& \frac{1}{\sqrt{N}} \|\mathcal{X}^\top\|_{2,1} = \frac{1}{\sqrt{N}} \|Q^\top \mathcal{V}^\top\|_{2,1} \text{,}
	\label{Eq_21}
\end{eqnarray}
where $\|.\|_{2,1}$ is the  $\ell_{2,1}$ norm of a matrix. According to Eq. (\ref{Eq_bigPi}) and Eq. (\ref{Eq_21}), we can simply maximise $\|Q^\top \mathcal{V}^\top\|_{2,1}$ to maximise $\Pi$ for the purpose of information diffusion. Finally, we can combine the diffusion regularisation with Eq. (\ref{f3_XVA+O(V)}) and Eq. (\ref{f2_L-DW}) to form the overall loss function. Such a function aims to minimise the reconstruction error from attributes to visual features, meanwhile preserve the data structure and enable the information to diffuse to all dimensions:
\begin{eqnarray}
	\label{Eq_Final_Loss}
	\min_{P,Q,\mathcal{V}}&  J = \|\mathcal{X}_s-\mathcal{V}Q\|^2_F+ \|\mathcal{V}-\mathcal{A}_s P\|^2_F +\lambda Tr(\mathcal{V}^\top L\mathcal{V})&  \nonumber \\
	&-\beta\|Q^\top \mathcal{V}^\top\|_{2,1}, ~~s.t.~ QQ^\top= I .&
\end{eqnarray}

\subsection{Optimisation Strategy}
The problem raised in Eq. (\ref{Eq_Final_Loss}) is a non-convex optimisation problem. To the best of our knowledge, there is no direct way to find its optimal solution. Similar to \cite{kodirov2015unsupervised}, in this paper, we propose an iterative scheme by using the alternating optimisation to obtain the local optimal solution. Specifically, we initialise $Q=I$ and $\mathcal{V}=\mathcal{X}_s$.The initialisation of $P$ can be obtained via $P=(\mathcal{A}_s^\top\mathcal{A}_s)^{-1}\mathcal{A}_s^\top\mathcal{V}$.  The whole alternate procedure of the proposed UVDS is listed as follows.

\noindent\textbf{1. $\mathcal{V}$-step:} By fixing $P$ and $Q$, we can reduce Eq. (\ref{Eq_Final_Loss}) to the following sub-problem:
\begin{align}
	\label{sub11}
	\min_{\mathcal{V}}& \|\mathcal{X}_s-\mathcal{V}Q\|^2_F+ \|\mathcal{V}-\mathcal{A}_s P\|^2_F +\lambda Tr(\mathcal{V}^\top L\mathcal{V})&  \nonumber \\
	&-\beta\|Q^\top \mathcal{V}^\top\|_{2,1}+ \gamma\|\textbf{1}\mathcal{V}\|_2^2,
\end{align}
where the extra term $\gamma\|\textbf{1}\mathcal{V}\|_2^2$ constrains the learnt $\mathcal{V}$ to be centralised according to Eq. \ref{theorem1_invariance}. The minimal $\mathcal{V}$ can be obtained by setting the  partial derivative of Eq. (\ref{sub11}) to zero and we have
\begin{eqnarray}
	\label{sub12}
	\frac{\partial J}{\partial \mathcal{V}}&=& 2(\mathcal{V}Q-\mathcal{X})Q^\top+2(\mathcal{V}-\mathcal{A}P) \nonumber\\
	&+& 2 \lambda L\mathcal{V} -\beta \mathcal{V}QEQ^\top +\gamma\textbf{1}^\top\textbf{1}\mathcal{V}= 0,
\end{eqnarray}
where $E = diag(e_1,\ldots,e_d,\ldots,e_D)\in\mathbb{R}^{D\times D}$ and the $d$-th element of $E$ is $e_d= 1/(\sqrt{N}\pi_d)$.
By merging the like terms, Eq. (\ref{sub12}) can be rewritten as
\begin{eqnarray}
	&\mathcal{V}(2QQ^\top+2\alpha I + \beta QEQ^\top)+ (2\lambda L+\gamma\textbf{1}^\top\textbf{1} )\mathcal{V} \nonumber\\
	&-(XQ^\top+2\mathcal{A}P)=0,
	\label{Eq_solveV}
\end{eqnarray}
which is  a typical Sylvester equation so that $\mathcal{V}$ can be efficiently solved by the \emph{lyap()} function in the MATLAB. Afterwards, the leant $\mathcal{V}$ needs to be further centralised: $v_n\leftarrow v_n- (\sum_{n=1}^{N} v_n )/N$ to satisfy Eq. \ref{theorem1_invariance}.\\

\noindent\textbf{2. Q-step}: By fixing $P$ and $V$, we can reduce Eq. (\ref{Eq_Final_Loss}) to the following sub-problem:
\begin{equation}
	\label{sub21}
	\min_{\mathcal{Q}} \|\mathcal{X}_s-\mathcal{V}Q\|^2_F-\beta\|Q^\top \mathcal{V}^\top\|_{2,1}, ~~s.t.~ QQ^\top= I
\end{equation}

Since we need to solve $Q$ with the orthogonality constraint in Eq. (\ref{sub21}), in this paper, we adopt the gradient flow in \cite{l21_solver} which is an iterative scheme for optimising generic orthogonal problems with a feasible solution.
Specifically, given the orthogonal rotation $Q_{t}$ during the $t$-th iterative optimisation, a better solution of $Q_{t+1}$ is updated via \textit{Cayley transformation}:
\begin{equation}
	\label{sub22}
	Q_{t+1}=H_{t}Q_{t},
\end{equation}
where $H_{t}$ is the \textit{Cayley transformation} matrix and defined as
\begin{equation}
	\label{sub23}
	H_{t}=(\mathrm{I}+\frac{\tau}{2} \Phi_{t} )^{-1}(\mathrm{I}-\frac{\tau}{2} \Phi_{t}),
\end{equation}
where $\mathrm{I}$ is the identity matrix, $\Phi_t = \Delta _t Q_{t}^\top-Q_{t}\Delta _t^\top$ is the skew-symmetric matrix, $\tau$ is an approximate minimiser satisfying Armijo-Wolfe conditions \cite{tau} and $\Delta$ is the partial derivative of Eq. (\ref{sub21}) with respect to $Q$ as
\begin{equation}
	\Delta_{t} = \mathcal{V}^\top(\mathcal{V}Q_{t}-\mathcal{X}_s)-\beta \mathcal{V}^\top\mathcal{V}Q_{t}E.
	\label{Eq_Qconv},
\end{equation}
where the diagonal matrix $E$ is defined the same as that in Eq. (\ref{sub12}). In this way, for the Q-step, we repeat the above formulation to update Q until achieving convergence.

\vspace{2ex}
\noindent\textbf{3. P-step}: By fixing $Q$ and $V$, we can reduce Eq. (\ref{Eq_Final_Loss}) to the following sub-problem:
\begin{equation}
	\label{sub31}
	\min_{P} \alpha\|\mathcal{V}-\mathcal{A}_s P\|^2_F.
\end{equation}
The resulted equation is derived by a standard least squares problem with the following analytical solution:
\begin{equation}
	P=(\mathcal{A}_s^\top\mathcal{A}_s)^{-1}\mathcal{A}_s^\top\mathcal{V}.
	\label{Eq_solveP}
\end{equation}

In this way, we sequentially update $\mathcal{V}$, $\mathbf{Q}$ and $\mathbf{P}$ to optimise UVDS with $T$ times based on coordinate descent. For each variable, either global or local optimum is achieved and thus the overall objective is lower bounded, which guarantees the convergence of our method.  In practice,  UVDS can well converge with $T=5\sim 10$.

%
\begin{algorithm}\label{A1}
	\caption{Unseen Visual Data Synthesis (UVDS)}
	\KwIn {Training set $\{\mathcal{X}_s,~\mathcal{A}_s,~\mathcal{Y}_s\}$, $k$ for $k$-nn graph}
	\KwOut{$P,~Q\text{ and } \mathcal{V}$}
	Initialise $Q=\mathrm{I}$, $\mathcal{V}=\mathcal{X}_s$ and $P=(\mathcal{A}_s^\top\mathcal{A}_s)^{-1}\mathcal{A}_s^\top\mathcal{V}$, where $\mathrm{I}\in\mathbb{R}^{D\times D}$ is the identity matrix.\\
	\textbf{Repeat}\\
	~~~\textbf{$\mathcal{V}$-Step:} Fix $P$, $Q$  and update $\mathcal{V}$ using Eq. (\ref{Eq_solveV}).\\
	~~~\textbf{$Q$-Step:} Fix $P$, $\mathcal{V}$  and update $\mathcal{Q}$ by following steps:\\
	~~~\textbf{for} $t=1: \mathrm{max\ iterations}$ \textbf{do}\\
	~~~~~~~Compute the gradient $\Delta_{t}$ using Eq. (\ref{Eq_Qconv});\\
	~~~~~~~Compute the the skew-symmetric matrix $\Phi_t$;\\
	~~~~~~~Compute the Cayley  matrix $H_{t}$ using Eq. (\ref{sub23});\\
	~~~~~~~Compute the $Q_{t+1}$ using Eq. (\ref{sub22});\\
	~~~~~~~\textbf{if} convergence, \textbf{break};\\
	~~~\textbf{end}\\
	~~~\textbf{$P$-Step:} Fix $\mathcal{V}$, $Q$  and update $\mathcal{P}$ using Eq. (\ref{Eq_solveP}).\\
	\textbf{Until} convergence\\
	\textbf{Return}  $f_{UVDS}(x)=xPQ$\\
\end{algorithm}


\begin{table*}[t]
	\centering
	\caption{Comparison with State-of-the-art methods. }
	\label{T2_comp_stat}
	\resizebox{0.98\textwidth}{!}{
		\begin{tabular}{l|c|llll}
			\hline\Xhline{1pt}
			Methods                    &Feature & Animals with Attributes        & Caltech-UCSD Birds        & aPascal\&aYahoo        & SUN Attribute       \\
			\hline
			DAP \cite{2014_AwA2_PAMI}                       &$\mathcal{L}$       & 40.50         & -          & 18.12             & 52.50      \\
			ALE \cite{2013_Label_Embedding}                       &$\mathcal{L}$       & 43.50         & 18.00      & -                 & -          \\
			Jayaraman and Grauman \cite{2014_Unreliable}      &$\mathcal{L}$       & 43.01$\pm$ 0.07 & -          & 26.02$\pm$ 0.05 & 56.18$\pm$ 0.27 \\
			Romera-Paredes and Torr \cite{2015_embarrassingly}   &$\mathcal{L}$       & 49.30$\pm$ 0.21 & -          & 27.27$\pm$ 1.62 & -          \\
			\hline
			Ours+CA                    &$\mathcal{L}$       & \textbf{53.45$\pm$ 0.30}& \textbf{43.52$\pm$ 0.69} & 36.98$\pm$ 0.62     & 53.46$\pm$ 1.32      \\
			Ours+SVM                   &$\mathcal{L}$       & -               & 40.88$\pm$ 1.34     & \textbf{44.21$\pm$ 0.28 }    & \textbf{66.03$\pm$ 0.74}      \\
			\Xhline{1pt}
			DAP \cite{2014_AwA2_PAMI}                         &$\mathcal{V}$     & 57.23      & -          & 38.16      & 72.00      \\
			Akata \cite{2015_akata_evaluation} &$\mathcal{A}$     & 61.9       & 40.3       & -          &  -         \\
			Romera-Paredes and Torr \cite{2015_embarrassingly}   &$\mathcal{V}$     & 75.32$\pm$ 2.28 & -               & 24.22$\pm$ 2.89 & 82.10$\pm$ 0.32 \\
			Zhang and Saligrama  \cite{2015_semantic_similarity}      &$\mathcal{V}+T$     & 76.33$\pm$ 0.83 & 30.41$\pm$ 0.20 & 46.23$\pm$ 0.53 & 82.50$\pm$ 1.32 \\
			Zhang and Saligrama  \cite{2016_ZZiming}      &$\mathcal{V}+T$     & 80.46$\pm$ 0.53 & 42.11$\pm$ 0.55 & 50.35$\pm$ 2.97 & 83.83$\pm$ 0.29 \\
			Zhang and Saligrama  \cite{2016_ECCV_ZZming}      &$\mathcal{V}+T$     & \textbf{90.25 $\pm$ 8.08} & \textbf{53.30$\pm$ 33.39} & \textbf{65.36$\pm$ 37.29} & 86.00$\pm$ 14.97\\
			\hline
			Ours+CA                    &$\mathcal{V}$     & 82.12$\pm$ 0.12     & 44.90$\pm$ 0.88     & 42.25$\pm$ 0.54     & 80.50$\pm$ 0.75       \\
			Ours+SVM                   &$\mathcal{V}$     & -                 & 45.72$\pm$ 1.23   & 53.21$\pm$ 0.62  & \textbf{86.50$\pm$ 1.75 } \\
			\hline
		\end{tabular}}\\
		$\mathcal{L}$: Low-level feature, $\mathcal{A}$: Deep feature using AlexNet, and $\mathcal{V}$: VGG-19, CA: class-level attributes. T: transductive.
	\end{table*}
	\subsection{Zero-shot Recognition}
	Once we obtain the embedding matrices $P$ and $Q$, the visual features of unseen classes can be easily synthesised from their semantic attributes:
	\begin{equation}
		\mathcal{X}_u=\mathcal{A}_u PQ\text{.}
	\end{equation}
	
	It is noticeable that for image-level attributes, $\mathcal{X}_u$ contains as many instances as the test set. The zero-shot recognition task now becomes a typical classification problem. Thus, any existing supervised classifier, \eg SVM, can be applied. For class-level, only a prototype feature of each class is synthesised. Either few-shot learning techniques or the simplest Nearest Neighbour (NN) algorithm can be adopted. Since we focus on the quality of the synthesised features, we simply use NN and SVM for image-level tasks and NN for class-level tasks.

	\section{Experiments}
	
	\noindent\textbf{Settings} We evaluate our method on four benchmark datasets and strictly follow the published seen/unseen splits. For AwA \cite{2009_AwA} and aPY \cite{2009_aPY}, we follow the standard 40/10 and 20/12 splits like most of existing methods. For CUB, we follow \cite{2013_Label_Embedding} to use the 150/50 setting. For SUN, we use the simple 707/10 setting as reported in \cite{2014_Unreliable,2015_embarrassingly,2015_semantic_similarity}. Methods under different settings \cite{2013_NIPS_transductive,2015_PAMI_GsG_Multiview,2016_synthesized_Classifiers,2016_ECCV-GZSL}, or using other various semantic information \cite{2011_relative,2013_Category_level,2016_akata_Strong_Supervision,2015_akata_evaluation} are not compared with.
	
	\noindent\textbf{Semantic Attributes} Existing attributes are divided into image-level and class-level. On CUB, aPY, and SUN datasets, image-level attributes are provided. Our approach can synthesise the visual features for all unseen instances. We compute class-level attributes by averaging the image-level attributes for each class. For the AwA dataset, only class-level attributes are provided.
	
	\noindent\textbf{Visual Features} For low-level visual features, we use those provided by the four datasets \cite{2009_AwA,2009_aPY,SUN,CUB}. For deep learning features, we adopt CNN features released by\cite{2015_semantic_similarity} for the four datasets using the VGG-19 model.
	
	\noindent\textbf{Implementation Parameters} Half of the data in each class in the training sets are used as the validation set. We use 10-fold cross-validation to obtain the optimal hyper-parameters $\lambda$ and $\beta$. $k$ is fixed to 10 for the $k$-nn graph.
	
	\begin{table*}
		\centering
		\caption{Comparison with baseline methods.}
		\label{T3_SelfComp}
		\resizebox{0.98\textwidth}{!}{
			\begin{tabular}{l|l|cc|cc|cc|cc|cc|cc}
				\Xhline{1pt}
				\multirow{2}{*}{Scenario}        & Dataset                  & \multicolumn{4}{c|}{\textbf{CUB}}                               & \multicolumn{4}{c|}{\textbf{SUN}}                               & \multicolumn{4}{c}{\textbf{aPY}}                               \\
				\cline{2-14}
				& Test Domain              & \multicolumn{2}{c|}{Seen} & \multicolumn{2}{c|}{Unseen} & \multicolumn{2}{c|}{Seen} & \multicolumn{2}{c|}{Unseen} & \multicolumn{2}{c|}{Seen} & \multicolumn{2}{c}{Unseen} \\
				\Xhline{1pt}
				\multirow{5}{*}{Prototype-based} & \textbf{Baseline}                 & CA          & MF         & CA           & MF          & CA          & MF         & CA           & MF          & CA          & MF         & CA           & MF          \\
				\cline{2-14}
				& Linear Regression      & 66.82  & 64.34  & 27.28  & 30.31  & 88.85  & 89.12  & 63.00  & 64.50  & 52.42  & 55.35  & 17.96  & 21.63  \\
				& GR-only ($\beta = 0$)   & 65.79  & 65.53  & 38.82  & 40.42  & 89.67  & 88.41  & 75.50  & 76.00  & 59.38  & 57.75  & 25.75  & 28.86  \\
				& DR-only ($\lambda = 0$) & 66.32  & 67.98  & 37.75  & 40.64  & 90.31  & 89.85  & 74.00  & 77.50  & 57.96  & 58.32  & 30.28  & 32.46  \\
				& Ours                     & \textbf{67.47}  & \textbf{68.43}  & \textbf{44.90}  & \textbf{44.90}  & \textbf{92.32}  & \textbf{89.88}  &\textbf{ 80.50}  & \textbf{78.50}  & \textbf{62.75}  & \textbf{64.88}  & \textbf{42.25}  & \textbf{41.97}       \\
				\Xhline{1pt}
				\multirow{5}{*}{Sample-based}    & \textbf{Baseline}                 & NN          & SVM        & NN           & SVM         & NN          & SVM        & NN           & SVM         & NN          & SVM        & NN           & SVM         \\
				\cline{2-14}
				& Linear Regression     & \textbf{64.57}& 67.44         & 22.36 & 26.57 & \textbf{90.79}  & 92.27 & 72.50 & 77.00 & 43.75 & 44.42 & 13.48 & 15.96 \\
				& GR-only ($\beta = 0$)  & 61.38         & 66.88         & 32.65 & 38.58 & 88.42           & 91.91 & 74.50 & 80.00 & 53.34 & 57.08 & 22.74 & 25.59 \\
				& DR-only ($\lambda = 0$) & 62.44         & 68.94         & 36.93 & 42.24 & 88.34           & 90.47 & 78.00 & 84.00 & \textbf{55.05} & 53.41 & 23.68 & 24.22 \\
				& Ours                  & 63.78         & \textbf{70.32}& \textbf{39.82} & \textbf{45.72} & 89.85           & \textbf{93.23} & \textbf{78.50} & \textbf{86.50} & 54.35 &\textbf{69.75} & \textbf{38.49} &\textbf{53.21} \\
				\hline
			\end{tabular}}\\
			
			\scriptsize CA: Class-level attributes, MF: Mean of synthesised features, GR: Graph regularisation, and DR: Diffusion regularisation. Best results are in bold.
			\vspace{-2ex}
		\end{table*}
		\begin{figure*}[t]
			\centering
			\includegraphics[width=0.98\textwidth]{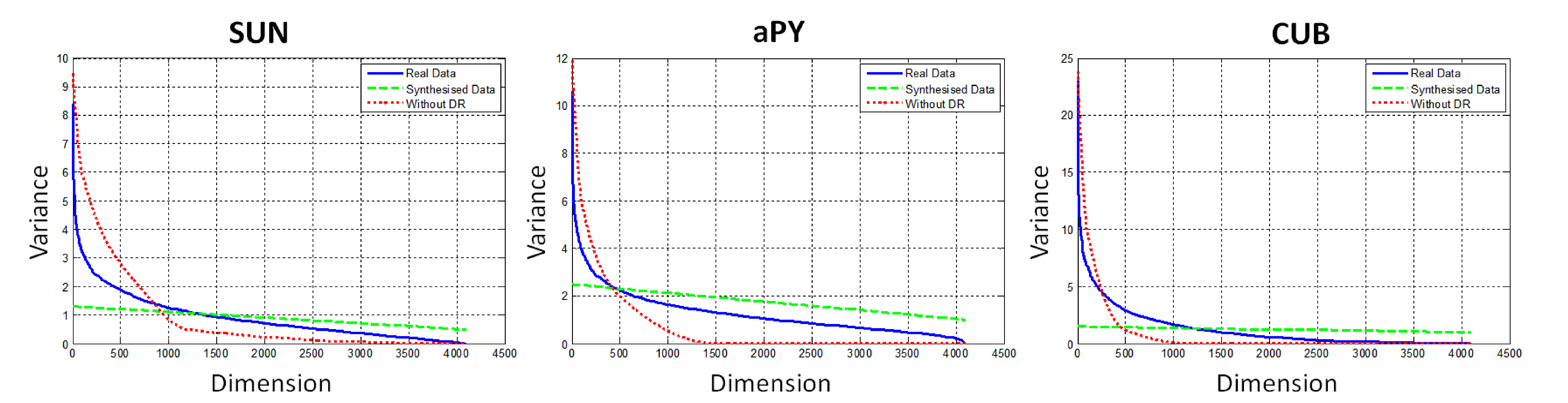}
			\caption{\small Normalised variances of the synthesised data \wrt dimensions. Variance of each dimension is sorted in descending order. We make a comparison between the synthesised data variances `with' (green) and `without' (red) diffusion regularisation. The variances of real data (blue) are computed from real unseen data as references.}
			\label{fig_DRcurve}
			\vspace{-2ex}
		\end{figure*}

		\subsection{Comparison with the State-of-the-art methods}
		Table \ref{T2_comp_stat} summarises our comparison to the published results of state-of-the-art methods. The hyphens indicate that the compared methods were not tested on the corresponding datasets in the original papers. In the first section, all of the compared methods were tested using conventional low-level features. In the second section, deep learning features are employed. For all of the four datasets, we first evaluate our method using class-level attributes (CA). In this scenario, each unseen class gains a synthesised visual feature prototype from the class attribute signature. The unseen test images are predicted by the NN classification using these prototypes. When image-level attributes are available in CUB, aPY, and SUN, we further conduct experiments using SVM classifiers. The visual feature vector of each unseen image is synthesised by the proposed UVDS and then fed to train SVM models. During the test, visual features that are extracted from the unseen images are fed to the trained SVM to get the prediction. Our method can steadily outperform the state-of-the-art methods on conventional ZSL scenarios. Our results also exceed two of the results base on transductive settings \cite{2016_ECCV_ZZming,2015_semantic_similarity}, which sufficiently support our synthesised visual features are highly discriminative. While deep learning features can boost the performance, our method can also achieve acceptable results with low-level features. In most cases, using SVM can further improve the recognition rates, especially when the class-level attributes are noisy, \eg on aPY and SUN. However, if the class-level attributes are more precise, \eg CUB, the class-level NN classifier can be better than SVM.  
		
		\subsection{Detailed Evaluations}
		\noindent\textbf{Baseline methods} To understand the effect of each term in Eq. (\ref{Eq_Final_Loss}), we compare our method to several baseline methods in Table \ref{T3_SelfComp}. Since AwA only provides class-level attributes, the following experiments are conducted on CUB, SUN, and aPY only. The first method is simply \textit{Linear Regression} that we solve Eq. (\ref{f2_A2X}) and synthesise prototypes of unseen classes using Eq. (\ref{f3_XWA}). The second and third methods are denoted as \textit{Graph-Regularisation (GR) only} ($\beta = 0$) and \textit{ Diffusion-Regularisation (DR) only} ($\lambda = 0$). For the training bias problem, we use the validation set to test the methods on seen classes. We also investigate ZSL under both class-level and image-level attributes scenarios. The first scenario is \textit{prototype-based}, \ie each unseen class gains only one visual prototype. We compare two possible ways to obtain the class-level visual prototype: 1) we compute the mean of image-level attributes in each class and use the averaged class-level attributes (CA) to synthesise one visual prototype for each class; 2) we first synthesise the visual features from the image-level attributes and use the mean of the features (MF) as the class prototype. During the test, we use NN classification to predict the label for the test image. The second scenario is \textit{sample-based}, \ie each unseen image has one unique attribute description. In this scenario, we fully synthesise all of the visual features of unseen classes and use them as training examples. We show how an advanced classifier, \eg SVM, can further boost the performance.

		\begin{figure*}
			\centering
			\includegraphics[width=16cm]{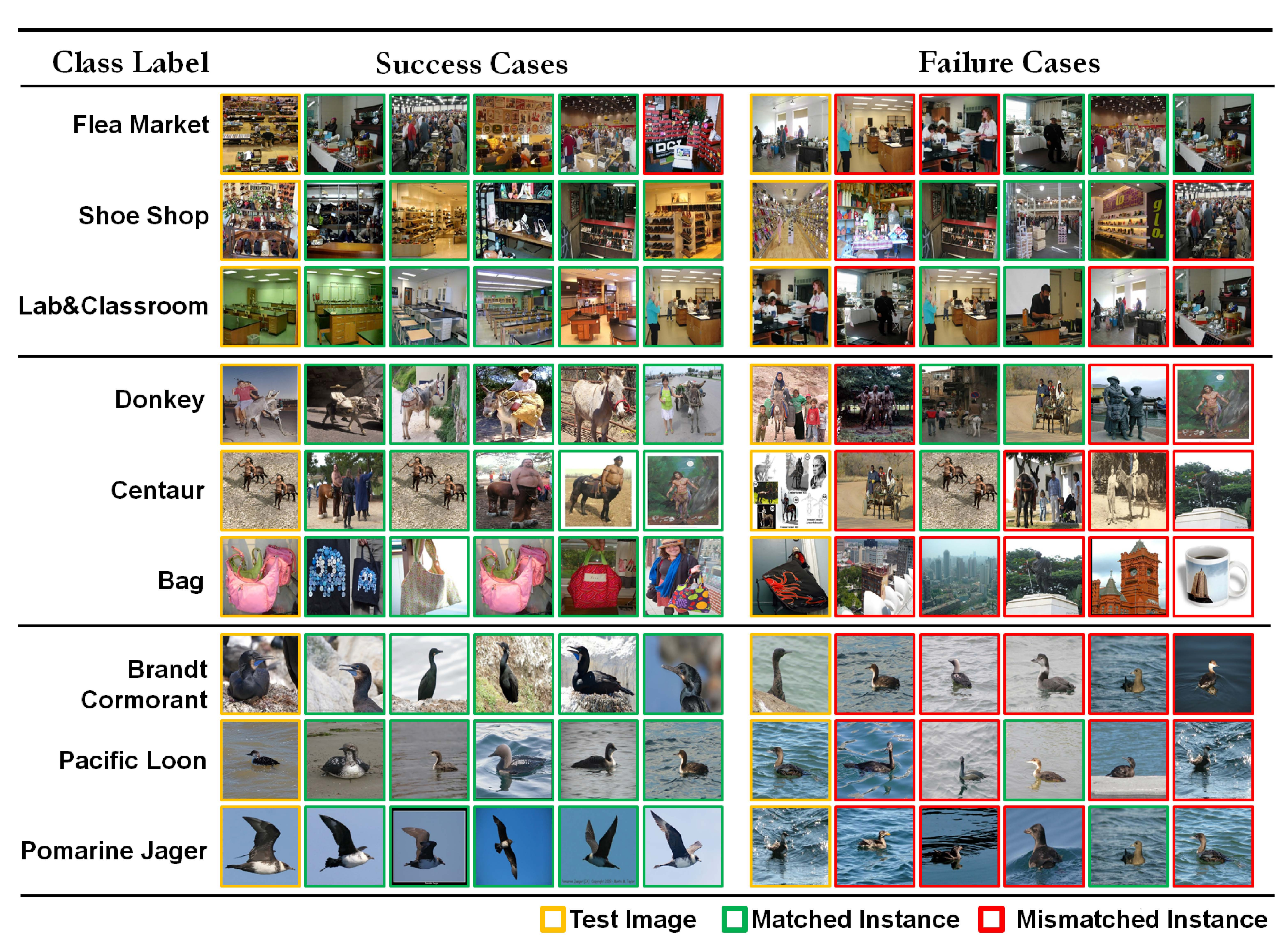}
			\caption{Success and Failure cases of nearest neighbour matching. The query visual feature is synthesised from its attribute description. We find top-5 nearest neighbours of the query feature from the real instances. It is a match if the nearest instance and the test image have the same label.}
			\label{fig_cases}
			\vspace{-2ex}
		\end{figure*}
		
		In summary, our method can effectively prevent the training bias whereas the linear regression without regularisation suffers from 30\% performance degradation in average from seen to unseen. DR is complementary to GR and can further boost the performance. There is no significant difference between the CA and MF scenarios. Therefore, our proposed method can be reliably applied to both image-level and class-level attributes. Another advantage is that the synthesised visual data can be fed to typical supervised classifiers to achieve better performance, which can be supported by the results using SVM.

		\noindent\textbf{Further Discussion} There are two more questions: (1) what are the outcomes of the diffusion regularisation? (2) What kind of visual features are synthesised? In Fig. \ref{fig_DRcurve}, we show the variance of each dimension of the synthesised data. The variances are sorted in descending order. We compare with the real unseen data and the synthesised data without diffusion regularisation ($\beta = 0$). Note that, in the synthesised data without DR (red), most variances are concentrated in a few dimensions (roughly 1000, 1500, and 500 on SUN, aPY, and CUB) while most of the remaining dimensions gain very low variances. In comparison, the variances of our proposed synthesised data (green) and real data are more informative. Furthermore, thanks to the DR, the variances in our proposed data are more balanced than real data, \ie each of the dimension gains the equal amount of information. Such quantitative evidence explains the success of our proposed method in ZSL recognition. 
		
		Finally, we provide some qualitative results of our method. We use the synthesised features as queries and retrieve real images from the unseen datasets. In Fig. \ref{fig_cases}, we show some success cases that most of the top-5 results are with the same class labels. Particularly, the third result of \textit{Bag} is the same paired image of the attributes that are used to synthesise the data. Such results demonstrate that the synthesised data is close to the samples from the same class in the feature space. On the contrary, we also provide some failure cases that the top-1 retrieval result is not with the same class label. Some of them are due to the ambiguity of the semantic meaning, \eg the \textit{flea market} has many similar attributes to the \textit{shoe shop}. Some other cases, \eg the CUB dataset, the real data of the birds are not distinctive to the other classes. Therefore, the NN-based retrieval gives a mixture of true-positives and false-positives. Such failures due to the ambiguity of the visual feature are not common cases. We can still achieve 45.72\% overall recognition rate on the CUB dataset.
		
		\section{Conclusion}
		In this paper, we proposed a novel algorithm that synthesises visual data for unseen classes using semantic attributes. From the experiments, we can see that directly embedding using regression-based models can lead to low recognition rates owing to three main problems, in terms of structural difference, training bias, and variance decay. In correspondence, we introduced a latent structure-preserving space with the diffusion regularisation. Our approach outperformed the state-of-the-art methods on all of the four benchmark datasets. For future work, a worthy attempt is to substitute the semantic attributes by automatic word vectors that are driven from the text. In this way, the cost of synthesising data can be further reduced.

{\small
\bibliographystyle{ieee}
\bibliography{egbib2}
}

\end{document}